\documentclass[11pt]{article}
\usepackage{coling2020}
\usepackage{times}
\usepackage{url}
\usepackage{latexsym}
\usepackage{graphicx}
\usepackage{amsmath}
\usepackage{dsfont}
\usepackage{multirow}

\colingfinalcopy % Uncomment this line for the final submission

\title{MZET: Memory Augmented Zero-Shot Fine-grained Named Entity Typing}

\author{
  Tao Zhang{$^{1}$}, Congying Xia{$^{1}$}\thanks{~ Corresponding Author}, Chun-Ta Lu{$^2$}, Philip S. Yu{$^1$}\\
  {$^1$University of Illinois at Chicago, Chicago, IL, USA;} \\
  {$^2$Google Research, Mountain view, CA, USA} \\
  {\tt \{tzhang90,cxia8,psyu\}@uic.edu; chunta@google.com}
}

\begin{document}
\maketitle
\begin{abstract}
Named entity typing (NET) is a classification task of assigning an entity mention in the context with given semantic types. However, with the growing size and granularity of the entity types, few previous researches concern with newly emerged entity types. In this paper, we propose MZET, a novel memory augmented FNET (Fine-grained NET) model, to tackle the unseen types in a zero-shot manner. MZET incorporates character-level, word-level, and contextural-level information to learn the entity mention representation. Besides, MZET considers the semantic meaning and the hierarchical structure into the entity type representation. Finally, through the memory component which models the relationship between the entity mention and the entity type, MZET transfers the knowledge from seen entity types to the zero-shot ones. Extensive experiments on three public datasets show the superior performance obtained by MZET, which surpasses the state-of-the-art FNET neural network models with up to 8\% gain in Micro-F1 and Macro-F1 score.
\end{abstract}

\section{Introduction}
\label{intro}
Named entity typing (NET) is the task of inferring semantic types for the given named entity mentions in utterances. For instance, given an entity mention ``John'' in the utterance ``John plays piano on the stage". The goal for NET is to infer that ``John" is a pianist or a musician, and a person. Standard NET approaches \cite{chinchor1997muc,sang2003introduction,doddington2004automatic} only consider a tiny set of coarse-grained types, and discard fine-grained types with a different level of granularity.
In recent years, fine-grained named entity typing (FNET) \cite{ling2012,naka2013,del2015,ren2016,anan2017,zhou2019} continues to draw researchers' attention, because it can provide additional information that benefits a lot of downstream tasks like relation extraction \cite{liu2014exploring}, entity linking \cite{stern2012joint}, and question answering \cite{han2017answer}.

However, with the ever-growing number of entity types especially for fine-grained ones, it is difficult and expensive to collect sufficient annotations per category and retrain the whole model. Therefore, a zero-shot paradigm is welcomed in FNET to handle the increasing number of unseen types. The task we deal with in this paper is named zero-shot fine-grained named entity typing (ZFNET), which is to detect the unseen fine-grained entity types that have no labeled data available. 

Learning generalizable representations for entity mentions and types is essential for the ZFNET task. Previous works learn these representations either from hand-crafted features \cite{ma2016,yuan2018}, or pre-trained word embeddings \cite{ren2016}. These methods are insufficient and inefficient when challenged by poly-semantic, ambiguity, or even the newly-emerged mentions. The most recent works \cite{obei2019,zhou2019} learn more informative but resource-costing representations by assembling the exterior Wikipedia knowledge base.

With the learned representations for entity mentions and entity types, most of the existing zero-shot FNET methods \cite{ma2016,obei2019} project them into a shared semantic space. 
The shared space is learned through minimizing the distance between entity mentions and its corresponding seen entity types.
In the prediction phase, testing entity mentions are classified to the nearest unseen entity types based on the assumption that the learned distance measurement also works for unseen types. These methods' ability to transfer knowledge from seen types to unseen types is limited since they do not explicitly build connections between seen types and unseen types. 

In this work, we propose the memory augmented zero-shot FNET model (MZET) to tackle the aforementioned problems. MZET is designed to automatically extract the multi-information integrated mention representations and structure-aware semantic type representations with a large-scale pre-trained language model \cite{devlin2018bert}. To effectively transfer knowledge from seen types to unseen types, MZET regards seen types as memory components and explicitly models the relationships between seen types and unseen types. Intuitively, we want to mimic the way how humans learn new concepts. Humans learn new concepts by comparing the similarities and differences between new concepts and old concepts stored in our memory.

In summary, the main contributions of MZET are as follows. 1) We propose the memory augmented zero-shot FNET model (MZET) that can be trained in an end-to-end fashion. MZET extracts multi-information integrated mention representations and structure-aware semantic type representations without additional augmented data sources. 2) MZET regards seen types as memory components and explicitly models the relationships between seen types and unseen types to effectively transfer knowledge to new concepts. 3) MZET outperforms existing zero-shot FNET models significantly on the zero-shot fine-grained, coarse-grained, and hybrid-grained named entity typing over three benchmark datasets.

\section{Problem Definition}
\label{sect:Problem}
We begin by formalizing the problem of zero-shot fine-grained named entity typing (ZFNET).
For a given entity mention $x$, the task of named entity typing (NET) is to identify the type $y$ for $x$. Suppose we have a training type set $\mathcal{Y}_{seen} = \{y_1^s, y_2^s, ..., y_{D_s}^s\}$ with $D_s$ seen types. There are a large number of labeled examples available for these seen types, $\mathcal D_{tr} = \{(x_i, y_i), i = 1, 2, ..., |\mathcal{D}_{tr}|\}$ with $y_i \in {\mathcal Y_{seen}}$.

The task of ZFNET is to classify a new mention which belongs to one of the unseen fine-grained entity types $\mathcal{Y}_{unseen} =  \{y_1^u, y_2^u, ..., y_{D_u}^u\}$, where $D_u$ is the number of unseen fine-grained entity types and $\mathcal{Y}_{seen} \cap \mathcal{Y}_{unseen} = \emptyset$.

\begin{figure*}[ht!]
    \centering
    \includegraphics[width= 0.9\textwidth]{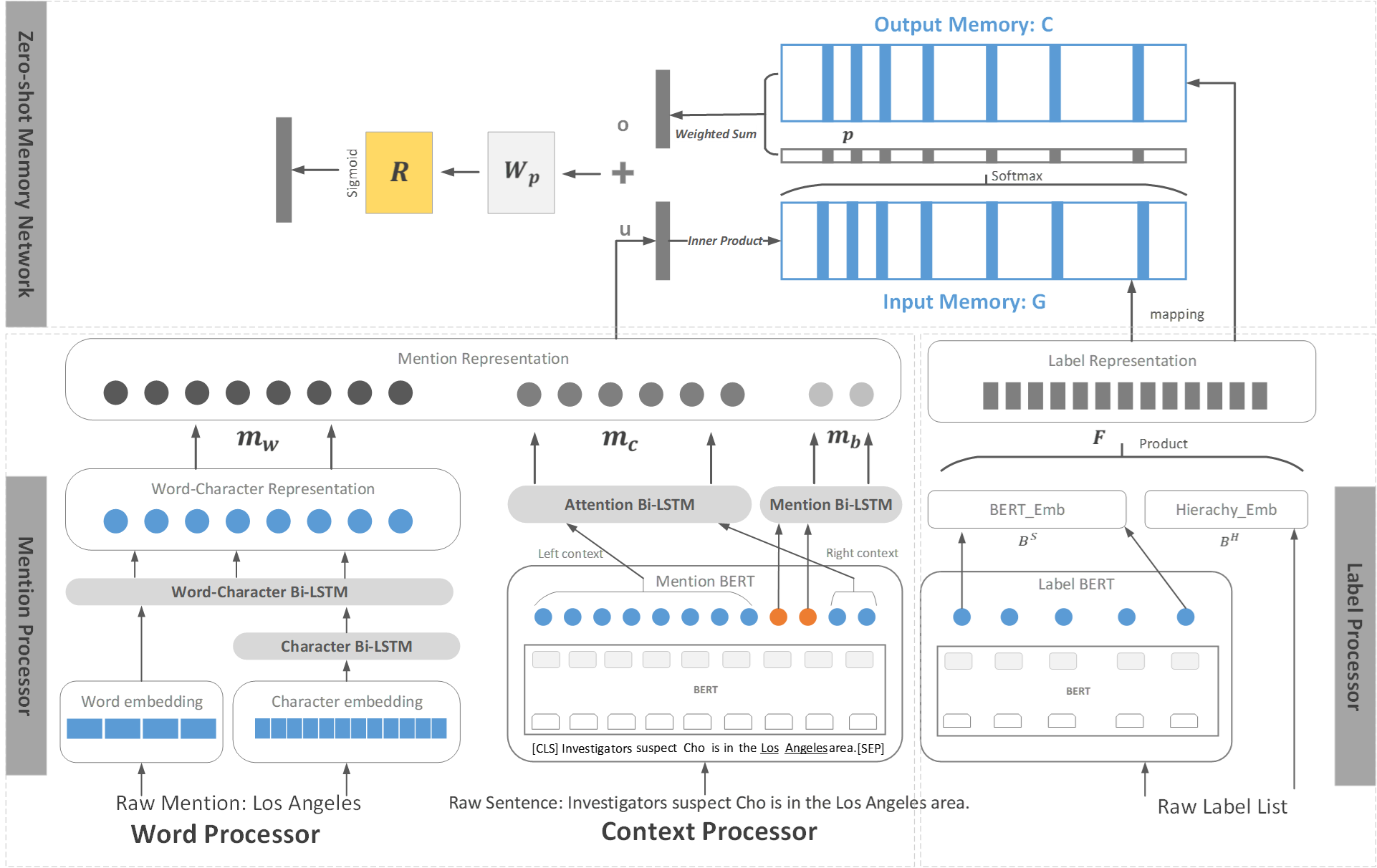}
    \vspace{-0.05in}
    \caption{The framework of MZET for zero-shot fine-grained named entity typing. It consists of three main components: mention processor, label processor, and zero-shot memory network.} 
    \vspace{-0.1in}
    \label{fig:framework}
\end{figure*}

\section{The proposed Model}
\label{sect:Model}
The overview of the proposed MZET framework is illustrated in Figure \ref{fig:framework}. Specifically, MZET consists of three components: 1) Zero-shot Memory Network that identifies entity types for entity mentions, introduced in Sec.\ref{ssect:Memory Network}; 2) Mention Processor which extracts representation for entity mentions, detailed in Sec.\ref{ssect:Mention Processor}; 3) Label Processor which obtains type representation, depicted in Sec.\ref{ssect:Label Processor}.

\subsection{Zero-Shot Memory Network}
\label{ssect:Memory Network}
In the zero-shot entity typing task, there are no mentions available for these unseen entity types. Without the labeled data, we are not able to model the direct mapping from the new mentions to the new types. Here, we propose a novel zero-shot memory network that utilizes seen entity types to bridge the gap between the new mentions and the zero-shot entity types.

\subsubsection{Memory augmented Typing Function}
\label{sssect:Memory augmented Typing Function}
To enable the zero-shot paradigm, previous researches \cite{ma2016,obei2019} introduce a score function $f(\cdot)$ to rate the match of a given entity mention $x$ and an entity type $y$, where $y$ is the raw type picked from $\mathcal{Y}_{seen}$ or $\mathcal{Y}_{unseen}$. The definition for $f(\cdot)$ is:
\vspace{-0.08in}
\begin{equation}
\begin{gathered} 
    f(x,y) = \theta(x,A)\cdot \phi(y, B) \\
    = Ax\cdot By,
\end{gathered}
\vspace{-0.08in}
\end{equation}
where $\theta(x,A): x\rightarrow Ax$ and $\phi(y, B): y\rightarrow By$ serve as the mapping functions that project $x$ and $y$ into a shared semantic space by neural networks (ours are depicted in Sec. \ref{ssect:Mention Processor} and Sec.\ref{ssect:Label Processor} respectively). $f(\cdot)$ is the distance estimation of $Ax$ and $By$ in the shared space. 

Considering the lack of interpretability of the representations in the shared semantic space, we propose the memory network augmented zero-shot FNET to construct another high-level shared space, called \textit{Association Space}. Each dimension in the association space links to an entity type. The representation in this space indicates the association information with each entity type. The score function for memory augmented zero-shot FNET is changed into $f'$:
\vspace{-0.08in}
\begin{equation}
\begin{gathered}
    f'(x,y) = MEM_{\mathcal{Y}_{seen}}( \theta(x,A),\phi(y, B)) \\
    = MEM_{\mathcal{Y}_{seen}}(Ax, By),
\end{gathered}
\vspace{-0.08in}
\end{equation}
where $MEM_{\mathcal{Y}_{seen}}(\cdot)$ means rating score estimated by a memory network for the zero-shot paradigm with $\mathcal{Y}_{seen}$ as the memory component. Meanwhile, the memory component is the aforementioned association space to guide unseen entity typing by linking them to each seen type stored in the memory components, like humans recognizing new things by intuitively associating with the knowledge they have memorized. In fact, $MEM_{\mathcal{Y}_{seen}}(\cdot)$ loads $Ax$ and $By$ in the shared semantic space into the high-level association space, and then estimates matching score under the help of their connections to the seen types in the memory.

\subsubsection{Zero-Shot Memory Network Model}
\label{sssect:Zero-Shot Memory Network Model}
All the seen entity representations are utilized as the memories in the zero-shot memory network. We propose to use the memory network as a special attention mechanism to model the relationships between the mentions and the seen entity types. Furthermore, we build a zero-shot version memory network that utilizes the type representation similarities to transfer the knowledge from the seen types to the unseen types, which exactly implement the detail of $MEM_{\mathcal{Y}_{seen}}(\cdot)$. The key points to implement $MEM_{\mathcal{Y}_{seen}}(\cdot)$ are interpreted as follows: (1) The \textit{Association Space} is constructed with all the seen types representations to bridge the gap between mentions and unseen types. (2) The mention representation is augmented by the association with seen types. That means to obtain attention between mentions and seen types as the association. After absorbing the association, the mentions obtain more informative representations benefiting the knowledge transferring in the \textit{Association Space}. (3) The augmented mention and type representation are projected into the \textit{Association Space}. Association augmented mention can be directly project into it. But for the unseen types, associations between them with the seen types are formed by the type semantic similarity, which exactly presents each unseen type in the \textit{Association Space}.

We first construct the \textit{Association Space} with all the seen types representations from the Label Processor (in Sec.\ref{ssect:Label Processor}), $\mathbf{F} = (\mathbf{f}^s_1, ..., \mathbf{f}^s_{D_s}) \in \mathds{R}^{D_s \times D_b}$,  where $D_s$ is the number of seen types and $D_b$ is the dimension of the type representation. 

To augment the mention representation by it association with seen types, we construct two dependant memory components $\mathbf{G}$ and $\mathbf{C}$.
As shown in Figure \ref{fig:framework}, the input memory representation $\mathbf{G} = (\mathbf{g}_1, ..., \mathbf{g}_{D_s}) \in \mathds{R}^{D_s \times D_m}$ is converted from $\mathbf{F}$ using an embedding matrix $\mathbf{W}_{f1} \in \mathds{R}^{D_b \times D_m}$, where $D_m$ is the dimension of the memory components. To catch the association between mentions and seen labels through memory component, we model the attention $p_i$ between the mention input $\mathbf{u}$ and each memory component $\mathbf{g}_i \in \mathds{R}^{D_m}, i \in \{1,...,D_s\}$ with:
\vspace{-0.08in}
\begin{equation}
p_i = \text{softmax}(\mathbf{u}^\top \mathbf{g}_i),
\vspace{-0.08in}
\end{equation}
where $\mathbf{u}$ = $\mathbf{W}^\top \mathbf{m}$, and $\mathbf{W} \in \mathds{R}^{D_e \times D_m}$. $\mathbf{m}$ is the mention representation that is obtained from the Mention Processor in Sec.\ref{ssect:Mention Processor}, and its dimension size is $D_e$. The input memory representation $\mathbf{G}$ works to update the association $\mathbf{P}$ for each mention $\mathbf{u}$. We construct the output memory representations $\mathbf{C} \in \mathds{R}^{D_s \times D_m}$ from $\mathbf{F}$ using another embedding matrix $\mathbf{W}_{f2} \in \mathds{R}^{D_b \times D_m}$. The attentions $p_i$ are used as weights to associate the output memory representations and obtain the associated mention embedding:
\vspace{-0.08in}
\begin{equation}
\mathbf{o} = \sum_i{p_i \mathbf{c}_i}.
\vspace{-0.08in}
\end{equation}
Finally, $(\mathbf{o} + \mathbf{u})$ is the adjusted mention representation augmented by the information associated with seen types. Then it is projected into the association space by $\mathbf{W_p} \in \mathds{R}^{D_s \times D_m}$.

To load unseen types into the association space for the zero-shot capability of our memory network, we use the similarities between the type representations to transfer knowledge from seen types to unseen types. The similarities between type $\mathbf{f}_i$ and $\mathbf{f}_j$ are calculated as:
\vspace{-0.08in}
\begin{equation}
r_{ij} = \frac{\text{exp}\{-d(\mathbf{f}_i, \mathbf{f}_j)\}}{\sum^{D_s}_{j=1}{\text{exp}\{-d(\mathbf{f}_i, \mathbf{f}_j)\}}},
\vspace{-0.04in}
\end{equation}
where $d(\mathbf{f}_i, \mathbf{f}_j)$ is the Euclidean distance between $\mathbf{f}_i$ and $\mathbf{f}_j$. $\mathbf{f}_i$ is from $\mathbf{F} = (\mathbf{f}^s_1, ..., \mathbf{f}^s_{D_s}) \in \mathds{R}^{D_s \times D_b}$. $\mathbf{f}_j$ is from $(\mathbf{f}^u_1, ..., \mathbf{f}^u_{D_u}) \in \mathds{R}^{D_u \times D_b}$ during zero-shot testing, while $\mathbf{f}_j$ is from $\mathbf{F}$ for training the model. Then we can get the similarity matrix $\mathbf{R} \in \mathds{R}^{D_s\times D_u}$ for all the unseen types during prediction. We use the associated mention embedding $\mathbf{o}$, the mention input $\mathbf{u}$, and the similarity matrix $\mathbf{R}$ together to classify the zero-shot entity types in the association space:
\vspace{-0.08in}
\begin{equation}
y = \text{sigmoid}(\mathbf{R}^\top \mathbf{W_p}(\mathbf{o} + \mathbf{u})).
\vspace{-0.04in}
\end{equation}

In this way, we construct a 2-level shared space for zero-shot FNET by the memory network as shown in Figure \ref{fig:framework}. The lower one is the semantic representation space, which is formed by $\mathbf{m}$ in Sec.\ref{ssect:Mention Processor} and $\mathbf{f}$ in Sec.\ref{ssect:Label Processor}. The higher one is the association space that models the connections between not only mentions and seen types but also seen types and unseen types. Therefore, we can tell the reasoning process of the prediction from the association space.  For instance, the association space contains seen type ``/SUBSTANCE", and ``/DRUG". Given the mention ``pills", it matches the unseen type ``/SUBSTANCE/DRUG", as both ``pills" and ``/SUBSTANCE/DRUG" associate with the seen type ``/SUBSTANCE", and ``/DRUG". 

We can also extend the memory components to handle multiple hop operations \cite{sukhbaatar2015end} by stacking the memories sequentially which leaves for the future work.

\subsection{Mention Processor}
\label{ssect:Mention Processor}
To better understand the entity mention, we not only consider the words contained in the mention, but also the context around it. The Mention Processor has two sub-components. A Word Processor is proposed to get the semantic meaning for each word in the entity mention. Another Context Processor is utilized to understand the sequential information together with the context. The final mention representation $\mathbf{m}$ is a concatenation of the word-level representation from the Word Processor and the sequential representation from the Context Processor as shown in Figure \ref{fig:framework}. 

\subsubsection{Word Processor}
\label{sssect:Word Processor}
Following most existing works \cite{lample2016neural,lin2018neural,bari2019zero}, Word Processor is proposed to achieve basic understandings over the words in the entity mentions. Given an input entity mention $\mathbf{X}_w=(t_1, ..., t_K)$ with $K$ tokens, each token $t_k$ is represented as $
\mathbf[\mathbf{w}_k;\mathbf{c}_k].$ It is a concatenation of a pre-trained word embedding $\mathbf{w}_k \in \mathds{R}^{D_w}$ ($D_w$ is the dimension of the pre-trained word embedding) and a character-level embedding $\mathbf{c}_k$ which provides morphological information and makes a complement when faced with out-of-vocabulary (OOV) words. 
The character-level embedding $\mathbf{c}_k\in \mathds{R}^{D_{h_c}}$ is obtained through a bi-directional LSTM ($D_{h_c}$ is the dimension size after concatenating bi-direction hidden states), named as Character Bi-LSTM. %Random initialized character embeddings are fed into the Character Bi-LSTM. The final hidden states from the forward and backward LSTM are concatenated as $\mathbf{c}_k$.

Additionally, another bi-directional LSTM, named as Word-Character Bi-LSTM, is utilized to gather the information from all the token embeddings $\mathbf{X}_w$ by concatenating the forward and backword hidden states, $\overrightarrow{\mathbf{h}}^w_t$ and $\overleftarrow{\mathbf{h}}^w_t$, respectively:%$\overrightarrow{\mathbf{h}}^w_t=\mathbf{G}^\mathbf{f}_{\theta_w}(\mathbf{X}_w,\overrightarrow{\mathbf{h}}^w_{t-1})$ and $\overleftarrow{\mathbf{h}}^w_t=\mathbf{G}^\mathbf{b}_{\theta_w}(\mathbf{X}_w,\overleftarrow{\mathbf{h}}^w_{t+1})$, respectively:
\vspace{-0.08in}
\begin{equation}
\mathbf{m}_{w} = \overrightarrow{\mathbf{h}}^w_t\oplus \overleftarrow{\mathbf{h}}^w_t,
\vspace{-0.04in}
\end{equation}
where $\mathbf{m}_{w}\in \mathds{R}^{D_{h}}$, $D_{h}$ is the dimension after concatenating Word-Character Bi-LSTM hidden states. As illustrated in Figure \ref{fig:framework}, the Word Processor outputs a word-character embedding $\mathbf{m}_w$ for each entity mention.

\subsubsection{Context Processor}
\label{sssect:Context Processor}
In the Context Processor, we leverage the powerful pre-trained language model, BERT \cite{devlin2018bert}, to incorporate two more context-aware parts into the mention representation: (1)$\mathbf{m}_{b}$, the mention embedding given the context; (2)$\mathbf{m}_{c}$, the surrounding context embedding.

%The intuition of $\mathbf{BERT}_m$ is BERT can ensure the mention embedding is compatible and coherent with its context by the multi-head attention mechanism. Meanwhile, the contextural embedding $\mathbf{Ctxt}_m$ is frequently adopted for information augment on the classification tasks.

Considering that a context-aware word embedding can carry syntax feature, we first conduct BERT to embed the whole sentence and obtain the BERT contextual embedding for each token. For the tokens contained in the entity mention, named as mention tokens, their BERT embeddings are represented as $\mathbf{X}_b=(\mathbf{b}_1, ..., \mathbf{b}_K)$, where $\mathbf{b}_i \in \mathds{R}^{D_b}$, and$D_b$ is the BERT embedding dimension. For the tokens in the surrounding context, named as context tokens, we only consider a fixed window for each mention to balance the computational cost. The BERT embeddings for left context tokens are $\mathbf{e}^l_1,..., \mathbf{e}^l_n$, and those in the right are $\mathbf{e}^r_1,..., \mathbf{e}^r_n$, where $\mathbf{e}^j_i \in \mathds{R}^{D_b}$ and $j\in\{l, r\}$.  $n$ is the window size and we set it as 10.
 
We utilize Bi-LSTMs (with concatenated bi-directional hidden state size $D_{h}$) to aggregate the separated token embeddings to extract the mention embedding $\mathbf{m}_b$ and the context embedding $\mathbf{m}_c$. $\mathbf{m}_b$ is obtained from the BERT embeddings of mention tokens $\mathbf{X}_b$ with the Bi-LSTM, called as Mention Bi-LSTM:
\vspace{-0.08in}
\begin{equation}
    \mathbf{m}_b=\overrightarrow{\mathbf{h}}^b_t\oplus \overleftarrow{\mathbf{h}}^b_t,
\vspace{-0.08in}
\end{equation}
%where $\mathbf{m}_b \in \mathds{R}^{D_{h}}$. The forward and backword hidden states are $\overrightarrow{\mathbf{h}}^b_t=\mathbf{G}^\mathbf{f}_{\theta_b}(\mathbf{X}_b,\overrightarrow{\mathbf{h}}^b_{t-1})$ and $\overleftarrow{\mathbf{h}}^b_t=\mathbf{G}^\mathbf{b}_{\theta_b}(\mathbf{X}_b,\overleftarrow{\mathbf{h}}^b_{t+1})$, respectively.
where $\mathbf{m}_b \in \mathds{R}^{D_{h}}$. $\overrightarrow{\mathbf{h}}^b_t$ and $\overleftarrow{\mathbf{h}}^b_t$ are the forward and backword hidden states of Mention Bi-LSTM, respectively.
$\mathbf{m}_c$ is obtained from the context tokens with a bi-directional LSTM with attention mechanism, called as Attention Bi-LSTM.
The hidden states in the bi-directional LSTM for the context tokens are denoted as:
$\overrightarrow{\mathbf{h}^l_1},\overleftarrow{\mathbf{h}^l_1}, ..., \overrightarrow{\mathbf{h}^l_n},\overleftarrow{\mathbf{h}^l_n}$, and 
$\overrightarrow{\mathbf{h}^r_1},\overleftarrow{\mathbf{h}^r_1}, ..., \overrightarrow{\mathbf{h}^r_n},\overleftarrow{\mathbf{h}^r_n}$. The attentions over all the context tokens are computed using a 2-layer feed forward neural network:
$
    \mathbf{e}^j_i = \text{tanh}(\mathbf{W}_e[
    \overrightarrow{\mathbf{h}^j_i};
    \overleftarrow{\mathbf{h}^j_i}
    ])$, $ 
    \tilde{\mathbf{a}}^j_i = \text{exp}(\mathbf{W}_a \mathbf{e}^j_i),
$
where $\mathbf{h}^j_i \in \mathds{R}^{D_h/2}$, $\mathbf{W}_e \in \mathds{R}^{D_h\times (D_a + D_a)}$, $\mathbf{W}_a \in \mathds{R}^{1 \times D_a}$, $D_a$ is the attention dimension, and $j \in \{l,r\}$. Then we normalize the attentions over all the mention tokens to obtain:
$
    \mathbf{a}^j_i = \frac{\tilde{\mathbf{a}}^j_i}{\sum^n_{i=1}{\tilde{\mathbf{a}}^l_i+\tilde{\mathbf{a}}^r_i}}.
$
The context embedding $\mathbf{m}_c \in \mathds{R}^{D_h}$ is weighted by the attentions:
\vspace{-0.08in}
\begin{equation}
    \mathbf{m}_c = \sum^n_{i=1}{(\mathbf{a}^l_i
    [
    \overrightarrow{\mathbf{h}^l_i};
    \overleftarrow{\mathbf{h}^l_i}
    ]
    +\mathbf{a}^r_i
    [
    \overrightarrow{\mathbf{h}^r_i};
    \overleftarrow{\mathbf{h}^r_i}
    ])}.
\vspace{-0.08in}
\end{equation}

\subsubsection{Mention Representation}
\label{sssect:Mention Representation}
The final entity mention representation with dimension $D_e \in \mathds{R}^{(D_h + D_h + D_h)}$ concatenates the word-character embedding, the mention embedding, and the context embedding as follow:
\vspace{-0.08in}
\begin{equation}
\mathbf{m} = [\mathbf{m}_w; \mathbf{m}_b; \mathbf{m}_c].
\vspace{-0.08in}
\end{equation}

\subsection{Label Processor}
\label{ssect:Label Processor}
Understanding the label is important in our task, since there is no information other than the label name for the zero-shot entity types. 
In the Label Processor, we get the semantic embeddings $B^S \in \mathds{R} ^{(D_s + D_u) \times D_b }$ for all the label names, including the seen labels ${\mathcal Y_{seen}}$ and the unseen labels $ {\mathcal Y_{unseen}}$, using a pre-trained BERT model.

The fine-grained labels and coarse-grained labels in ${\mathcal Y_{seen}}$ and $ {\mathcal Y_{unseen}}$ consist a hierarchical structure naturally. Each fine-grained type includes a coarse-grained type as the root in the hierarchical structure.  Following \cite{ma2016}, we utilize a sparse matrix $B^H \in \mathds{R} ^{\left(D_s + D_u\right) \times \left(D_s + D_u\right) }$ to represent the hierarchical structure in the labels. Each row $B^H_i$ corresponds to a binary hierarchical embedding for label $y_i$. For each entry in $B^H_i$, we use 1 to denote the label itself and its parent node, 0 for the rest:
\vspace{-0.08in}
\begin{equation}
    B^H_{ij} = \begin{cases} 1, &\text{if } { i=j} \text{ or }  {y_j \in Parent(y_i);}\\
    0, &\text{otherwise.}
    \end{cases}
\vspace{-0.08in}
\end{equation}

In the Label Processor, we integrate the semantic embeddings of the child label and its parent label into a single embedding vector as the fine-grained label representftion.
For a label $y_i$, the final label representation $\mathbf{f} \in \mathds{R}^{D_b}$ is represented together by the semantic embedding $B^S$ and its hierarchical embedding $B^H_i$ as shown in Figure \ref{fig:framework}:
\vspace{-0.08in}
\begin{equation}
\mathbf{f} = {B^S}^\top {B^H_i}.
\vspace{-0.08in}
\end{equation}

\subsection{Loss function}
\label{ssect:Loss function}
We train our model with a multi-label max-margin ranking objective as follows:
\vspace{-0.08in}
\begin{equation}
\mathcal{L} = \sum_{pos \in Y}\sum_{neg \in \overline{Y}} max(0, 1-p_{pos}+ p_{neg}).
\vspace{-0.08in}
\end{equation}
Given example mention $x$, $Y$ is the set of correct types assigned to $x$, $p_{pos}$ is the possibility for such a positive assignment. In contrast, $\overline{Y}$ is the set of incorrect assigned types. $p_{neg}$ is the possibility to assign a false label $neg\in \overline{Y}$ to $x$.

\section{Experiments}
\label{sect:Experiments}

\subsection{Datasets}
\label{ssect:Datasets}
We evaluate the performance of our model on three public datasets that are widely used in FNET task.

\vspace{-0.08in}
\paragraph{BBN} \cite{weischedel2005bbn} consists of 2,311 WSJ articles that are manually annotated using 93 types in a 2-level hierarchy.
\vspace{-0.08in}
\paragraph{OntoNote} \cite{weischedel2011ontonotes} has 13,109 news documents where 77 test documents are manually annotated using 89 types in a 3-level hierarchy.
\vspace{-0.08in}
\paragraph{Wiki} \cite{ling2012} consists of 1.5M sentences sampled from 780k Wikipedia articles. 434 news sentences are manually annotated for evaluation. 112 entity types are organized into a 2-level hierarchy.

\subsection{Zero-shot Setting}
\label{ssect:Zero-shot setting}
We follow Ma \shortcite{ma2016} and Obei \shortcite{obei2019} to apply the zero-shot setting that the training set only contains coarse-grained types (level-1), while all fine-grained types (level-2) only appear in the testing data. For the OntoNotes dataset that has 3 levels, we combine the level-1 and level-2 as the coarse-grained typing for training, and level-3 as the fine-grained types for testing.
%So, we let the training set only contain coarse-grained types (level-1), while the testing set includes all fine-grained types (level-2). Especially, from Table \ref{Statistic}, we can see that OntoNotes only possesses 4 level-1 types. Hence, we combine the level-1 and level-2 as the coarse-grained typing for training, and level-3 as the fine-grained types for testing.

\subsection{Baselines}
\label{ssect:Baselines}
We compare the proposed method (MZET) and its variants with state-of-the-art FNET neural models. However few research approaches zero-shot FNET without auxiliary resource or hand-crafted features. In such a situation, we select the benchmarks and baselines as follows: 

\vspace{0.04in}
\noindent \textbf{DZET} Obei et al. \shortcite{obei2019} propose a neural structure to extract the mention representations but leverage Wikipedia to augment the label representations.
So we only compare with them on the learned mention representation capability, and incorporate our label embedding methods to construct this baseline.
%Unfortunately, their code and pre-processed dataset are not publicly available, so we re-implement their neural architecture as described in the paper, and evaluate it on our datasets.

\vspace{0.04in}
\noindent \textbf{OTyper} Yuan et al. \shortcite{yuan2018} devise a neural model for FNET, but still utilize pre-prepared hand-crafted mention features which are unavailable online. Furthermore, it is designed for open entity typing, which means to train and test the model on different datasets. Considered most parts of the model are learnable, we employ its results from their neural networks for comparison.

\vspace{0.04in}
\noindent \textbf{ProtoZET} Ma et al. \shortcite{ma2016} fist adapt zero-shot learning on FNET with hand-crafted features and propose prototype embedding to form label representation. Unfortunately, to the best of our knowledge, their system is not available online. We adopt its prototype label embedding technique and incorporate it with our Mention Processor for empirical comparison like Obei \shortcite{obei2019} and Zhou \shortcite{zhou2019} did before.

\vspace{0.04in}
\noindent \textbf{MZET + avg\_emb} For better contrast of label embedding techniques, we replace BERT label embedding in MZET with label average GloVe embedding that is widely adopted for entity typing in previous works \cite{shim2016,anan2017,yuan2018}.

%Unfortunately, to the best of our knowledge, the system proposed by Ma \shortcite{ma2016} is not available online for empirical comparison. But considering its contribution on label embedding method, we adopt its prototype label embedding to construct one of our variants (MZET+proto) like Obei \shortcite{obei2019} did before.

\subsection{Training and Implementation Details}
\label{ssect:Training & Implementation Details}
To train the neural network models, we optimize the multi-label max-margin loss function over training data concerning all model parameters. We adopt the Adam optimization algorithm with a decreasing learning rate of 0.0001, and the decay rate of 0.9. We utilize the pre-trained BERT (BERT-base, cased) with the number of transformer blocks is 12, the hidden layer size is 768, and the number of self-attention heads is 12. We also choose GloVe pre-training embeddings of size 300 for word-character representation. The hidden state of LSTMs is in size of 200.

We use hyperparameter $\tau$ as the maximum gap for selected labels. $\tau$ is optimized through validation sets (10\% of testing examples). Another strategy is for the prediction on overall dataset. We consider type inference over the predicted fine-grained type to include its parent coarse type into the final decision. Because we expect that such type inference can improve the recall score.

\subsection{Evaluation Metrics}
\label{ssect:Evaluation Metrics}
Following prior FNET works \cite{ling2012,ma2016,obei2019}, we evaluate our methods and baselines on three metrics: strict accuracy (Acc), Marco-F1, and Micro-F1. Given a collection of mention $M$, we denote the set of the ground truth and predicted labels of a mention $m\in M$ as $Y_m$ and $\hat{Y}_m$, respectively.
Strict Accuracy (Acc) = $\frac{\sum_{m\in M}\sigma{(Y_m = \hat{Y}_m)}}{M}$, where $\sigma(\cdot)$ is an indicator function.
 Macro-F1 is based on Macro-Precision($P_{ma}$) and Macro-Recall($R_{ma}$), where 
$
P_{ma}=\frac{1}{|M|}\sum_{m\in M}\frac{|Y_m \cap \hat{Y}_m|}{\hat{Y}_m}
$, 
$
R_{ma}=\frac{1}{|M|}\sum_{m\in M}\frac{|Y_m \cap \hat{Y}_m|}{Y_m}
$
 Micro-F1 is based on Micro-Precision($P_{mi}$) and Micro-Recall($R_{mi}$), where 
$
P_{mi}=\frac{\sum_{m\in M}|Y_m \cap \hat{Y}_m|}{\sum_{m\in M}\hat{Y}_m}
$, 
$
R_{mi}=\frac{\sum_{m\in M}|Y_m \cap \hat{Y}_m|}{\sum_{m\in M}Y_m}.
$

\vspace{-0.1in}
\begin{table*}[ht!]
\begin{center}
\begin{tabular}{l|lll|lll|lll}
\hline
\multicolumn{1}{c|}{\multirow{2}{*}{\textbf{Methods}}}           & \multicolumn{3}{c|}{\textbf{Overall}} & \multicolumn{3}{c|}{\textbf{Level 1}} & \multicolumn{3}{c}{\textbf{Level 2}} \\
\multicolumn{1}{c|}{}                                            & Acc       & Ma-F1       & Mi-F1      & Acc       & Ma-F1       & Mi-F1      & Acc       & Ma-F1       & Mi-F1      \\\hline
OTyper           & 0.203          & 0.447          & 0.451      & 0.501   & 0.589       & 0.591      & 0.189     & 0.202       & 0.209       \\
DZET + bert           & 0.214          & 0.481          & 0.509      & 0.517   & 0.634       & 0.665      & 0.207     & 0.234       & 0.246       \\
ProtoZET        & 0.251     & 0.582            & 0.631            & 0.620   & 0.676       & 0.677      & 0.214      & 0.239        & 0.257              \\
MZET + avg\_emb    & 0.285     & 0.588            & 0.669            & 0.672   & 0.691       & 0.691      & 0.262      & 0.293        & 0.304           \\
\textbf{MZET}    & \textbf{0.294}           & \textbf{0.606}            & \textbf{0.687}           & \textbf{0.700}            & \textbf{0.710}            & \textbf{0.710}            & \textbf{0.288}         & \textbf{0.301}            & \textbf{0.316}  \\ 
                                                \hline
\end{tabular}
\end{center}
\vspace{-0.05in}
\caption{\label{ZSL_FNET} Fine-grained entity typing evaluation on BBN dataset. DZET+bert utilizes mention embedding form Obei \shortcite{obei2019} and label embedding from BERT.}
\vspace{-0.05in}
\end{table*}

\begin{table*}[ht!]
\begin{center}
\begin{tabular}{l|lll|lll|lll}
\hline
\multicolumn{1}{c|}{\multirow{2}{*}{\textbf{Methods}}} & \multicolumn{3}{c|}{\textbf{BBN}}                & \multicolumn{3}{c|}{\textbf{OntoNotes}}          &  \multicolumn{3}{c}{\textbf{Wiki}}                                        \\ 
\multicolumn{1}{c|}{}                                & Acc            & Ma-F1          & Mi-F1          & Acc            & Ma-F1          & Mi-F1          & Acc                     & Ma-F1                   & Mi-F1                   \\ 
\hline
OTyper          & 0.211          & 0.497          & 0.513          & 0.211          & 0.256          & 0.259          & 0.269                   & 0.543                   & 0.547                   \\ 
DZET+bert                       & 0.236          & 0.530          & 0.542         & 0.231          & 0.276          & 0.281          & 0.285                   & 0.551                   & 0.560                   \\ 
ProtoZET                        & 0.251     & 0.582            & 0.631         & 0.281          & 0.337          & 0.345          & 0.296                   & 0.551                   & 0.564                   \\ 
MZET + avg\_emb        & 0.285     & 0.588            & 0.669          & 0.328          & 0.411          & 0.413          & 0.317                   & 0.554                   & 0.577                   \\ 

\textbf{MZET}                     & \textbf{0.294}           & \textbf{0.606}            & \textbf{0.687}  & \textbf{0.337} & \textbf{0.423} & \textbf{0.437} & \textbf{0.319}          & \textbf{0.555}          & \textbf{0.579}          \\ \hline
\end{tabular}
\end{center}
\vspace{-0.05in}
\caption{\label{3_Bench} The oeverall performance on 3 benchmark datasets. Results are evaluations of overall types including both coarse-grained and fine-grained types in the testing dataset.}
\vspace{-0.05in}
\end{table*}

\begin{table*}[ht!]
\begin{center}
\begin{tabular}{l|lll|lll|lll}
\hline
\multicolumn{1}{c|}{\multirow{2}{*}{\textbf{Methods}}}           & \multicolumn{3}{c|}{\textbf{Overall}} & \multicolumn{3}{c|}{\textbf{Level 1}} & \multicolumn{3}{c}{\textbf{Level 2}} \\
\multicolumn{1}{c|}{}                                            & Acc       & Ma-F1       & Mi-F1      & Acc       & Ma-F1       & Mi-F1      & Acc       & Ma-F1       & Mi-F1      \\\hline
\textbf{MZET}    & \textbf{0.294}           & \textbf{0.606}            & \textbf{0.687}           & \textbf{0.700}            & \textbf{0.710}            & \textbf{0.710}            & \textbf{0.288}         & \textbf{0.301}            & \textbf{0.316}  \\ 
MZET - Memory        & - 3.4          & - 2.2            & - 3.2           & - 1.1          & - 1.3            &- 1.3            & - 2.6         & - 2.3            &- 2.7  \\ 
MZET - Cntxt\_Attn      & - 2.0     & - 1.7            & - 2.1           & - 1.0   & - 1.3      & - 1.3      &
- 1.1      & - 1.4        & - 1.6      \\
MZET - Word\_Char        & - 2.7    & - 2.4            & - 2.8           & - 2.0   & - 2.2       & - 2.2      &
- 2.5      &- 2.0        & - 2.4      \\
MZET - BERT$_m$        & - 2.4    & - 2.1            & - 2.6           & - 1.8   & - 1.7       & - 1.7      &
- 2.5      &- 1.8        & - 2.5      \\
\hline
\end{tabular}
\end{center}
\vspace{-0.08in}
\caption{\label{Ablation} Ablation study on BBN dataset. All the results are percentages. The minus number means performance drop after remove or replace the methods. (-Memory) means replacing the memory part (\ref{ssect:Memory Network}) with regular zero-shot mapping function like the way in \cite{ma2016}. (-Cntxt\_Attn) means removing context representation $\textbf{m}_c$ in \ref{sssect:Context Processor}. (-Word\_Char) means removing word and character representation $\textbf{m}_w$ in \ref{sssect:Word Processor}. (-BERT$_m$) means remove mention BERT contextual embedding $\textbf{m}_b$ (\ref{sssect:Context Processor}).}
\vspace{-0.1in}
\end{table*}

\subsection{Results and Discussion}
\label{ssect:Results & Discussion}
\paragraph{Zero-Shot FNET Evaluation}
We first evaluate our methods for FNET on BBN dataset. 
Following  Ma\shortcite{ma2016}, we train the models on coarse-grained types, while testing in three ways:
%Zero-shot setting is that we keep the coarse-grained type for training the models, while testing in 3 ways: 
(1) Overall, predicting on both coarse-grained and fine-grained testing types; (2) Level 1, predicting only on coarse-grained types; (3) Level 2, predicting only on fine-grained types which are unseen before. Level-1 shows the performance for seen types, Level-2 evaluates the ability for zero-shot FNET, and Overall balances the performance between seen types and unseen types.

Table \ref{ZSL_FNET} illustrates the performance of the baselines and MZET on these 3 aspects. We see that for the coarse-grained typing (Level 1), MZET improvements strict accuracy significantly up to 19\%. For the zero-shot setting that testing on fine-grained types (Level 2), MZET achieves the highest scores and gains up to 10\% on strict accuracy. MZET attains the best with a 9\% gain on accuracy over the overall types. Compared to MZET+avg\_emb and ProtoZET, MZET gains significance performance from the Label Processor. Apart from the benefit from Label Processor, MZET also takes advantage of Mention Processor and Memory Network to achieve the best performer over the other baselines. At last, performance on all-grained types indicates the superiority of MZET over the rest, especially for the Micro-F1, which indicates the achievements over infrequent types.

To show the effectiveness of our proposed model, not only on the unseen fine-grained types but also on seen coarse-grained types, we evaluate the overall performance for three benchmark FNET datasets: BBN, OntoNotes, and Wiki. 
As Table \ref{3_Bench} shows, there are significant improvements of MZET on small datasets, BBN and OntoNotes. For the large-size Wiki data, MZET also attains the highest scores for all metrics. Compared with ProtoZET and MZET+avg\_emb, MZET shows only small improvement on Wiki but surpasses OTyper and DZET+bert almost 5\% on strict accuracy. This indicates that when the size of data increases, Mention Processor and Memory Network plays a great role for our model's ever-growing strength. As mentioned before, OntoNotes contains fine-grained entity type in 3-level hierarchy. Form the results over OntoNotes, MZET shows its superiority of Memory Network and Mention Processor with a significant margin especially for the most fine-grained dataset.

\vspace{0.08in}
\noindent \textbf{Ablation Study}
We carry out ablation studies that quantify the contribution of each component in our framework shown in Figure \ref{fig:framework}. As Table \ref{Ablation} shows, the vital parts are the memory network and the word and character representation. The performance decreases significantly over 2.5\% in strict accuracy by removing either of them. The memory network contributes decent augmentations on fine-grained typing, which indicates the noteworthy associations between the seen labels and mentions, as well as seen labels and unseen labels.  The word and character representation shows its importance on capturing the morphological and semantic information for a single entity mention. The secondary important is the informative context part with attention. It is aggregated into the final representation of the mention to guide the classification. Last, $\textbf{m}_b$ in Figure \ref{fig:framework} plays a considerable complementary role, as leading the BERT to embed a mention enables the model to gather more contextual information to avoid ambiguity for the polysemantic, like the word ``valley" in mention ``Silicon Valley". 

\begin{figure*}[ht!]
    \centering
    \includegraphics[width= 0.9\textwidth]{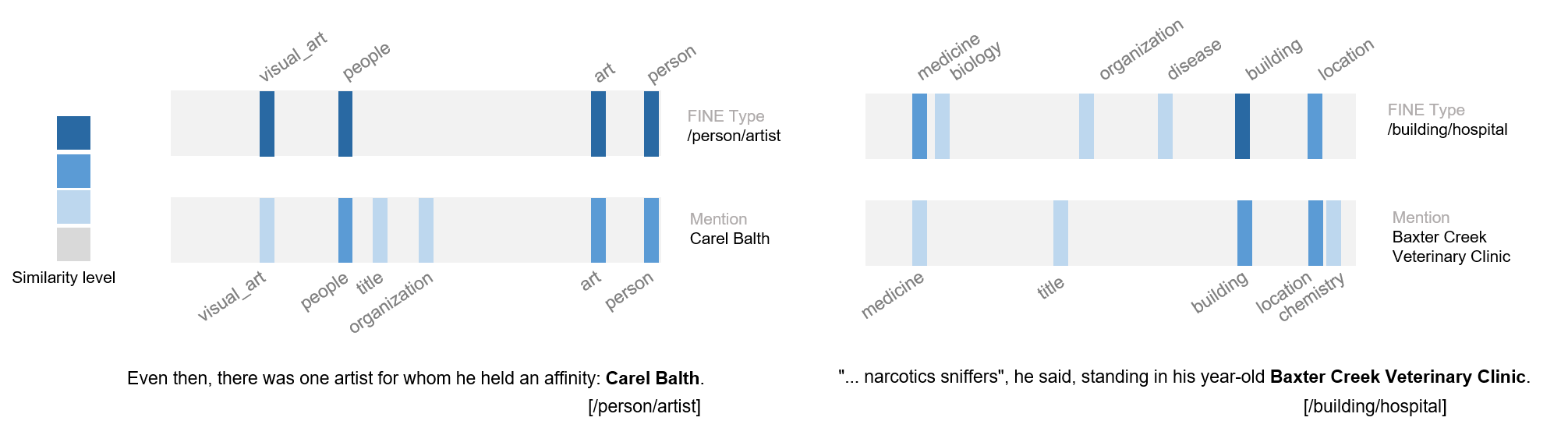}
    \vspace{-0.05in}
    \caption{Two examples to show the association between the zero-shot fine-grained type and a mention in an utterance. Their similarities with seen types are shown with the heat maps. Each dimension in the map links to one seen type. The above type heat map denotes the similarity between this unseen fine-grained type and all seen types. The mention heat map in the bottom is the output from Zero-shot Memory Network before the component $Sigmoid+\mathbf{R}$ in Figure \ref{fig:framework}.}
    \vspace{-0.1in}
    \label{fig:CaseStudy}
\end{figure*}

\vspace{0.08in}
\noindent \textbf{Case Study}
We visualize how to match the entity mention and type in the association space in Figure \ref{fig:CaseStudy}. Example 1 is a simple case as the unseen label words appear in seen types. This case shows agreement matching on most of the dimensions between the unseen type ``/person/artist" and the mention ``Carel Balth". Example 2 is more complex in the type similarity map as the new word ``hospital" has scattered associations with multiple seen types, like ``medicine'' and ``disease''. But they provide informative association about the unseen type for linking it to related mentions, like ``Baxter Creek Veterinary Clinic'' in the case.

\vspace{0.08in}
\noindent \textbf{Error Analysis}
We also provide insights into specific reasons for the mistakes made by our model. First, all the datasets follow long-tail frequency distributions. The examples for each label are significantly imbalanced. Accordingly, the model is prone to assign frequent types for the infrequent ones. For example, the training set processes 719 examples of ``/LOCATION" and 6,672 examples of ``/GPE" (Geopolitical Entity). The model prefers predicting on the fine-grained type ``/GPE/CITY"  rather than ``/LOCATION/REGION". 

Second, types are incorrectly tagged in the raw data. To test the ratio for incorrect tagging, we randomly pick out 100 examples in the raw data, including types coming from both training and testing sets. We find there are about $11\%$ for BBN, $10\%$ for OntoNotes, $13\%$ for Wiki with noise, such as mentions with incoherent labels, or missing the correct mention words for the corresponding tagged labels. For example, ``\emph{\underline{\textbf{The}} government estimates corn output at 7.45 billion bushels , up 51\% from last fall.}" labeled ``\emph{The}'' with two types: [``/ORGANIZATION/CORPORATION", ``/ORGANIZATION"]. The correct mention should be the ``\emph{The government}" other than ``\emph{The}'' for the assigned labels.

\vspace{-0.08in}

\section{Related Work}

FNET is a long-standing task in Natural Language Processing \cite{xia2019multi}. Most of the proposed FNET methods are based on a distant supervisor, but diverse in classification architectures. Ling et al.\cite{ling2012} propose multi-label and multi-class multilayer perceptron model assigns each mention of the corresponding label tags. Naka et al.\cite{naka2013} type newly emerging out-of Knowledge Base entities by a fine-grained typing system and harnesses relational paraphrase with type signatures for probabilistic weight computation. Del et al.\cite{del2015} designs a system, FINET, with the help from WordNet \cite{miller1995wordnet}. Ren et al.\cite{ren2016} propose AFET for automatic fine-grained entity typing with hand-crafted features and label embedding from the hierarchical type path. Shim et al.\cite{shim2016} and Anan et al.\cite{anan2017} adopt attentive neural network models for FNET. Attention information and contextural embedding are proposed to enhance FNET performance. Those methods develope, from hand-crafted features to neural network learned features, to allow fine-grained typing system fancy, automatic and effective. But their architectures can not apply to new and unseen entity types.

To handle unseen types, zero-shot learning \cite{xia2018zero} is introduced for named entity typing. Several works \cite{zhou2019,huang2016}  propose to solve unseen entity typing with clustering. These works cluster mentions and propagate type information from representative mentions to unseen types. Another direction is to construct a shared space for linking the seen and unseen data. 
These models \cite{ma2016,yuan2018,obei2019} map the mention and label embedding into a shared latent space, then estimate the closeness score for each mention-label pair. Most of the existing zero-shot FNET methods limit the model's flexibility with considerable auxiliary resources or pre-prepared hand-crafted features. In the perspective of entity type representations,  most recent researchers \cite{huang2016,zhou2019,obei2019} obtain informative entity type representations by assembling related Wikipedia pages. Their performances are decent yet resource-costing. Others \cite{ren2016,ma2016,shim2016,anan2017,yuan2018} exploit typical pre-trained semantic label embedding, which is easily-applied but pale in performance. Apart from various methods for entity type representation, mention reion approaches are also evolving recently.
Yuan \shortcite{yuan2018} and Ma \shortcite{ma2016} ultilize the pre-prepared hand-crafted features, while others \cite{obei2019,zhou2019} embed the mention by some pre-trained wording embedding methods \cite{2014glove,ELMo}. But, these methods are insufficient and inefficient when challenged by polysemantic, ambiguity, or even the newly-emerged mention.

\section{Conclusions}
\label{sect:Conclusions}
In this paper, we propose an end-to-end neural network, MZET, that enables zero-shot fine-grained named entity typing. It extracts comprehensive representations concerning word and character, mention, mention's context, and raw label text without auxiliary information. It adopts the memory network to gather the representations for zero-shot paradigm. Extensive experiments on three
public datasets show prominent performances obtained by MZET, which surpasses the state-of-the-art neural network models for Zero-Shot FNET.

%The contribution of this work is three-fold: First, such a novel neural network model can handle zero-shot FNET problems based on the information of the raw data without the assistant of additional augment resources. Second, we incorporate memory networks to indicate the connections mentions and labels and enable the zero-shot paradigm. Third, the robust performance of MZET attests to its contribution to the zero-shot FNET task.

\section{Acknowledgments}
We thank the reviewers for their valuable comments. This work is supported in part by NSF under grants III-1763325, III-1909323, and SaTC-1930941.

\bibliographystyle{coling}
\bibliography{coling2020}

\end{document}